\documentclass[letterpaper, 10pt, conference]{ieeeconf}      

\IEEEoverridecommandlockouts                              

\usepackage{cite}
\usepackage{graphicx}
\usepackage{xcolor, soul}
\usepackage{multirow}
\usepackage{array}
\usepackage{booktabs}
\usepackage{tikz}
\usepackage{makecell}
\usepackage{subcaption}
\usepackage[hidelinks=true]{hyperref}
\usepackage{amsmath}
\usepackage{amssymb}
\usepackage{siunitx}
\usepackage{fancyhdr}

\title{\LARGE \bf
Printed helicoids with embedded air channels make\\sensorized segments for soft continuum robots
}

\author{
  Annan Zhang, Hanna Matusik, Miguel Flores-Acton, Emily R. Sologuren, Joshua Jacob, Daniela Rus
\thanks{
  All authors are with the MIT Computer Science \& Artificial Intelligence Laboratory, Massachusetts Institute of Technology, 32 Vassar St, Cambridge, MA 02139, USA.
}
\thanks{Email correspondence to: \href{mailto:zhang@csail.mit.edu}{\tt\small zhang@csail.mit.edu}}
}

\begin{document}

\maketitle

\thispagestyle{fancy}
\fancyhf{} 
\renewcommand{\headrulewidth}{0pt} 
\fancyhead[C]{\footnotesize Accepted for publication in the proceedings of the \textit{2026 IEEE 9th International Conference on Soft Robotics (RoboSoft)}}
\fancyfoot[C]{\footnotesize \textcopyright 2026 IEEE. Personal use of this material is permitted. Permission from IEEE must be obtained for all other uses, in any current or future media, including reprinting/republishing this material for advertising or promotional purposes, creating new collective works, for resale or redistribution to servers or lists, or reuse of any copyrighted component of this work in other works.
}

\pagestyle{empty}

\begin{abstract}
Soft robots enable safe, adaptive interaction with complex environments but remain difficult to sense and control due to their highly deformable structures. Architected soft materials such as helicoid lattices offer tunable stiffness and strength but are challenging to instrument because of their sparse geometry. We introduce a fabrication method for embedding air channels into helicoid-based soft continuum robots. Multi-material segments fabricated via vision-controlled jetting in a single print interface with PCBs housing miniature pressure sensors and IMUs for distributed deformation sensing. We characterize the mechanical properties of four helicoid designs and validate the sensor response to fundamental deformation modes. To demonstrate the platform's scalability, we construct and mechanically evaluate a meter-scale, 14-DoF cable-driven soft arm capable of open-loop trajectory tracking and object grasping, with tactile-based stiffness detection demonstrated using the gripper sensors. This approach establishes a scalable fabrication strategy for sensorized architected materials in large-scale soft robotic systems.
\end{abstract}

\section{Introduction}

Soft robotics has emerged as a transformative paradigm for enabling safe, adaptive, and compliant interaction between robots and their environments~\cite{rus2015design,kim2013soft,majidi2014soft}. Unlike their rigid counterparts, soft robots can deform continuously, conform to complex surfaces, and safely manipulate delicate objects, which makes them ideal for applications ranging from biomedical devices to exploration and manufacturing. However, their deformability also makes modeling, actuation, and sensing profoundly challenging~\cite{yasa2023overview}. Robust control requires feedback on shape, force, and contact state, yet many current sensing methods fail on highly deformable, nonuniform, or sparsely structured materials~\cite{della2023model,wang2022control}.

Architected soft materials, mechanical metamaterials whose geometry defines their behavior, offer a promising direction for soft robotics~\cite{rafsanjani2019programming, guan2025lattice, mohammadi2023bioinspired, goswami20193d}. By tailoring lattice structures, these materials combine mechanical robustness with lightweight, tunable performance. Helicoid structures in particular have demonstrated mechanical efficiency and tunable directional compliance, which makes them well-suited for modular, cable-driven soft continuum robots~\cite{guan2023trimmed, patterson2025design}. Yet, a critical limitation persists: the lack of integrated sensing mechanisms capable of operating within these sparse, deformable, and geometrically complex structures.

Here, we introduce a new sensing approach for helicoid-based soft continuum robots using fluidic innervation. Building on recent advances in fluidic sensing~\cite{truby2022fluidic, zhang2024embedded, chen2024real, zhang2025fluidically, scharff2021sensing}, we develop a fabrication pipeline that integrates air channels directly within multi-material, 3D-printed helicoid segments using vision-controlled jetting~\cite{buchner2023vision}. The channels connect to custom PCBs with miniature pressure sensors and IMUs for multimodal deformation feedback. This enables distributed sensing without compromising mechanical integrity or scalability, and represents the first meter-scale helicoid-based continuum robot fabricated using vision-controlled jetting with embedded sensing.

\begin{figure}
  \centering
  \includegraphics[width = 1.\columnwidth]{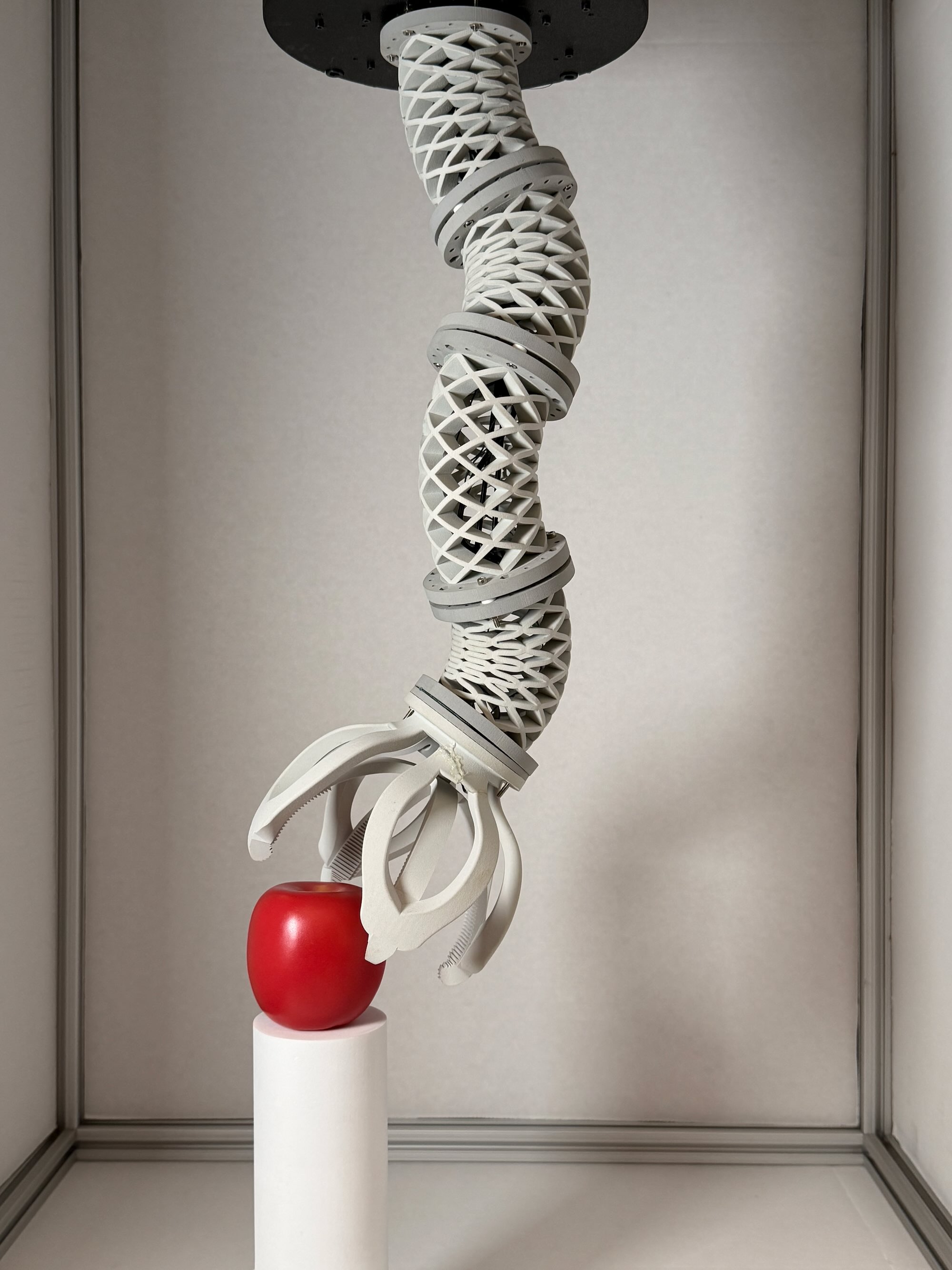}
  \caption{Overview of the multi-material soft continuum robot made of helicoid segments with embedded air channels.}
  \label{fig:overview}
  \vspace{-4mm}
\end{figure}

\subsection{Related Work}
Soft sensing technologies can be broadly categorized into piezoresistive, capacitive, optical, triboelectric, liquid metal-based, and vision-based systems. Piezoresistive and capacitive sensors are widely used for their simplicity.
However, piezoresistive sensors often suffer from poor repeatability, while capacitive sensors require compensation for parasitic capacitance~\cite{ji2020artificial,yousef2011tactile, peng2021recent}.
Triboelectric tactile sensors rely on triboelectric nanogenerators, but their limited output voltage restricts their real-time applications in continuous feedback systems~\cite{liu2025triboelectric, choi2023recent, lin2023self}.
Vision-based sensors, such as GelSight, have gained popularity due to their high resolution and ability to capture detailed information~\cite{yuan2017gelsight,azulay2023allsight,quan2022hivtac,do2022densetact,do2023densetact,sun2022soft,ward2018tactip,zhang2022vision,zhang2018robot,roberge2023stereotac,tippur2023gelsight360,gomes2020geltip}. However, due to the high computational demands of image processing, many vision-based systems operate at low frame rates~\cite{sun2022soft,ward2018tactip,roberge2023stereotac}. Moreover, these sensors often require elaborate lighting and optical configurations, which complicates their adaptability to different environments and geometries~\cite{tippur2023gelsight360,roberge2023stereotac}. Finally, the gel-like soft materials used in these systems degrade significantly over time~\cite{do2022densetact}.
Liquid metal-based sensors provide greater flexibility, but concerns related to leakage and material toxicity limit their practical applications~\cite{wang20213d,ren2020advances,qin2024emerging}. Optoelectronic sensors offer high precision but require complex setups of waveguides and light sources~\cite{zhao2016optoelectronically}.

These challenges are even more pronounced for robots with sparse architected material construction. The limited real estate and the fact that the entire wall structure is moving make it difficult to attach or embed sensors. Piezoresistive and capacitive sensors require mounting surfaces that remain relatively stable, but in sparse lattices every surface undergoes continuous deformation. Vision-based sensors require controlled optical environments with specific lighting configurations, but the open lattice structure allows ambient light penetration from all directions, which makes it difficult to maintain the necessary optical conditions. Liquid metal sensors would require complex routing through the moving lattice with heightened risk of leakage at the numerous structural vertices. In short, all of the aforementioned sensing methods face fundamental incompatibilities with sparse architected structures. This creates a critical sensing gap for architected soft robots and motivates the need for a sensing approach that works \textit{with} rather than against the sparse geometry.

\begin{figure}
  \centering
  \includegraphics[width = .72\columnwidth]{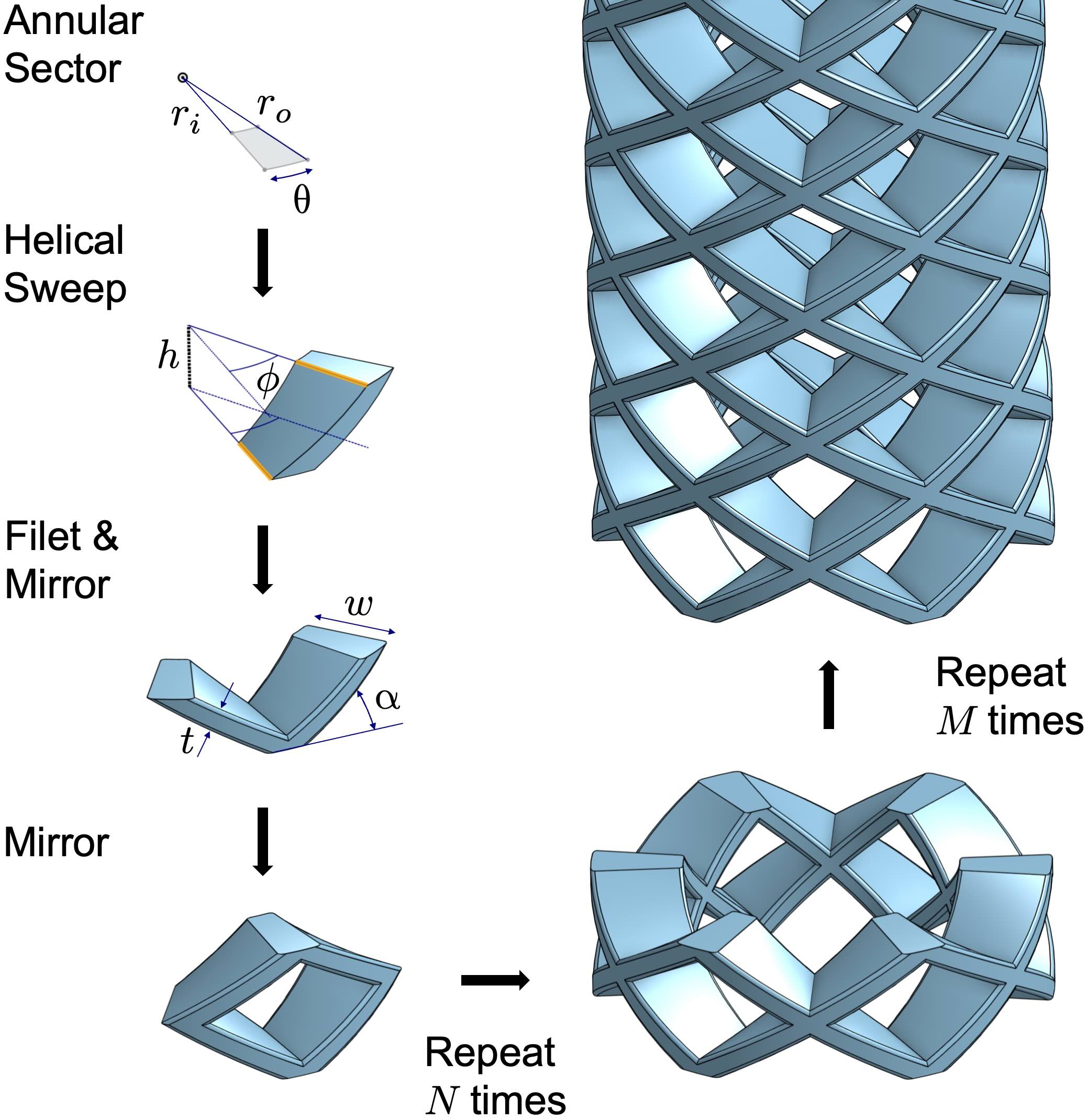}
  \caption{Helicoid construction showing annular sector parameters, helical sweep with axial rise and twist, and repeated structure.}
  \label{fig:helicoid_geometry}
  \vspace{-4mm}
\end{figure}

\subsection{Contributions}
The key contributions of this work are:
\begin{itemize}
    \item A fabrication method for multi-material, cable-driven continuum robot segments, 3D printed in one step using vision-controlled jetting;
    \item A sensing method for sparse helicoid-structured segments based on fluidic innervation;
    \item Mechanical characterization and analysis of four helicoid designs;
    \item The first meter-scale sensorized robot arm fabricated using vision-controlled jetting;
    \item Demonstration of the platform's mechanical capabilities through simple trajectory tracking and object manipulation;
    \item Proof-of-concept tactile sensing for object stiffness detection using gripper-embedded sensors.
\end{itemize}

\section{Hardware Design}
\label{sec:design}

The design of the proposed soft continuum robot leverages helicoid-based architected soft structures to achieve tunable stiffness and lightweight construction. To enable sensing without compromising the open architecture, fluidic channels are embedded along the helicoid walls and connected to pressure sensors for deformation feedback.

\subsection{Helicoid Design}

\begin{table}[t!]
\centering
\caption{Four helicoid designs under investigation}
\label{tab:parameters}
\begin{tabular}{lcccc}
\toprule
\textbf{Parameter} & \textbf{N4} & \textbf{N4T} & \textbf{N6} & \textbf{N8} \\
\midrule
Helicoid number, $N$ & 4 & 4 & 6 & 8 \\
Sector angle, $\theta$ ($^\circ$) & 18 & 24 & 12 & 9 \\
Twist angle, $\phi$ ($^\circ$) & 45 & 45 & 30 & 22.5 \\
Strut thickness at $r_i$, $t(r_i)$ (mm) & 3.04 & 4.05 & 2.60 & 2.22 \\
Strut thickness at $r_o$, $t(r_o)$ (mm) & 3.36 & 4.48 & 3.12 & 2.86 \\
\bottomrule
\end{tabular}
\end{table}

\begin{figure}[t!]
  \centering
  \includegraphics[width = .7\columnwidth]{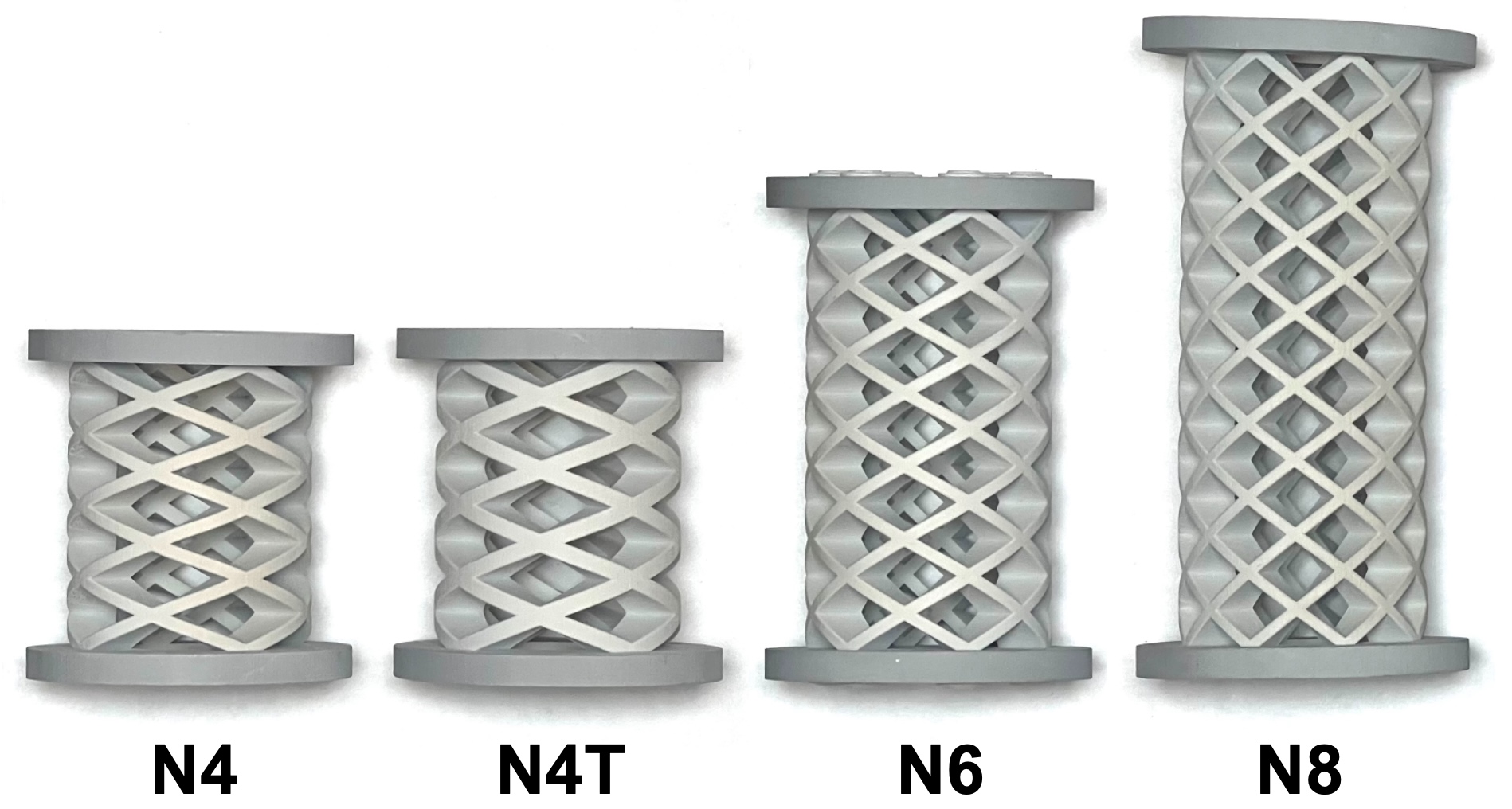}
  \caption{Comparison of the four helicoid designs showing the variation in strut thickness and spacing.}
  \label{fig:four_designs}
  \vspace{-4mm}
\end{figure}

The helicoid design (Fig.~\ref{fig:helicoid_geometry}) starts with a sector of a circular annulus with inner radius $r_i$, outer radius $r_o$, and sector angle $\theta$. This annular sector is moved up along the central axis by axial rise $h$ and rotated about the central axis by twist angle $\phi$. This helical sweep of the annular sector results in a volume that is a helical wedge of a thick cylindrical shell. This volume is mirrored twice and repeated $N$ times in the radial and $M$ times in the axial direction to form the helicoid structure.

To explore the mechanical design space and understand the tradeoffs between stiffness, compliance, and structural integrity, we fabricated and characterized four helicoid designs with varying helicoid number $N$ and sector angle $\theta$. All designs share common geometric parameters: inner radius $r_i = 18$ mm, outer radius $r_o = 30$ mm, radial width $w = 12$ mm, and axial rise $h = 9$ mm. The number of layers $M$ is chosen equal to $N$ to ensure that each of the $N$ helicoids makes a full 360$^\circ$ revolution over the segment height.

\begin{figure}
  \centering
  \includegraphics[width = 1.\columnwidth]{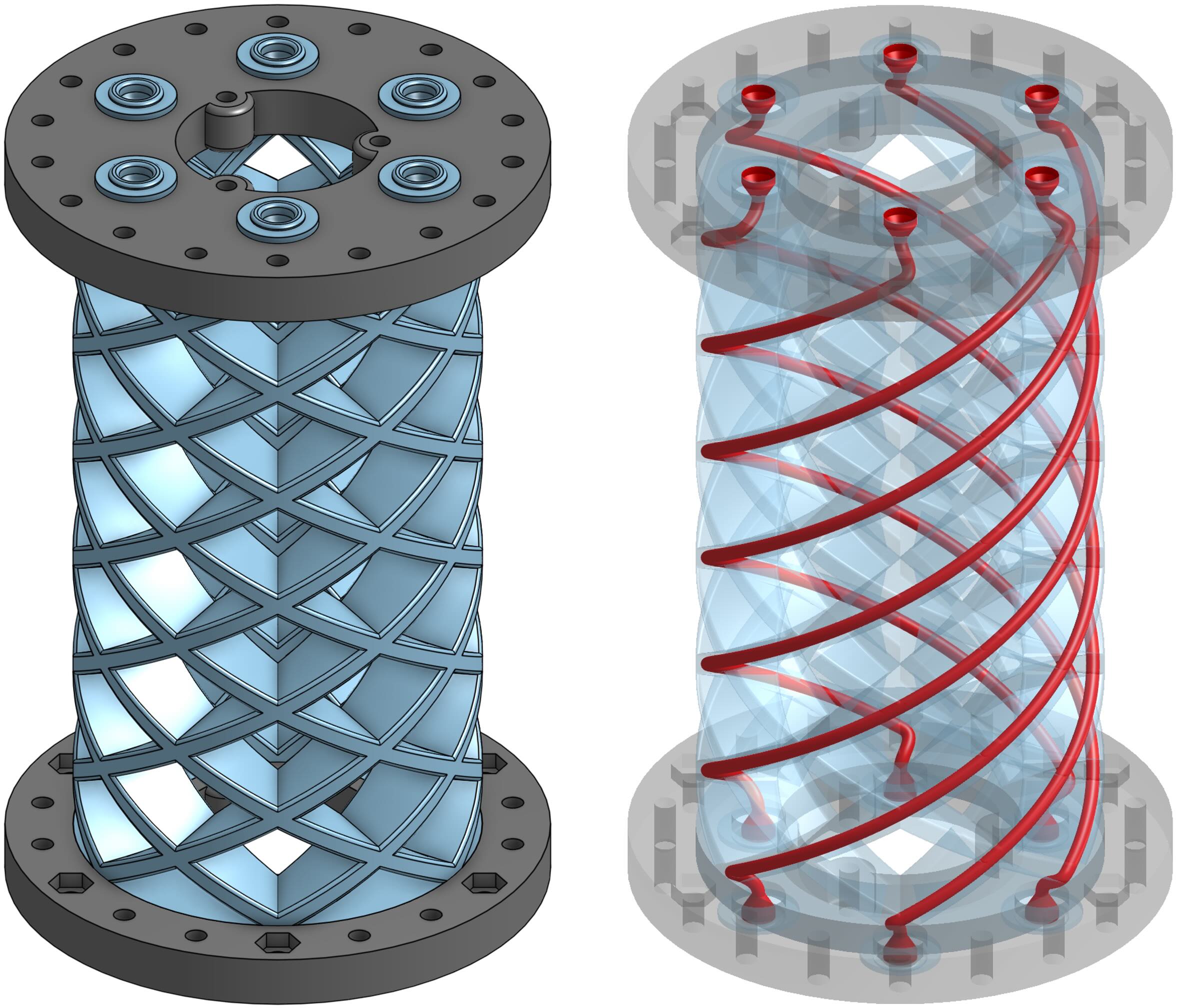}
  \caption{Left: N6 segment with rigid end plates and air channels. Right: Transparent view showing air channel routing through the helicoid structure.}
  \label{fig:air_channels}
  \vspace{-4mm}
\end{figure}

The design variants explore two key dimensions of the design space. First, varying $N$ (4, 6, 8) changes the number of helical struts and thus the structural redundancy and stiffness. Second, the N4T design uses a larger sector angle ($\theta = 24^\circ$) compared to N4 ($\theta = 18^\circ$), resulting in thicker struts while maintaining the same helicoid number. This allows us to decouple the effects of strut count versus strut thickness on mechanical performance.

From the base parameters, the twist angle follows as:
\begin{equation}
\phi = \frac{2\pi}{2N} = \frac{180^\circ}{N}
\end{equation}

The helix angle $\alpha$ (the angle between the tangent to the helix and a plane perpendicular to the central axis) varies with the radial coordinate $r$:
\begin{equation}
  \label{eq:alpha}
  \alpha(r) = \arctan\left(\frac{h}{r\phi}\right)
\end{equation}

The strut thickness also varies with radial coordinate $r$ and is given by the arc length of the annular sector at radius $r$ projected along the helix:
\begin{equation}
  \label{eq:t}
  t(r) = r \theta \sin(\alpha(r))
\end{equation}

Substituting Eq.~\ref{eq:alpha} and simplifying with the identity $\sin(\arctan(x)) = \frac{x}{\sqrt{x^2+1}}$ gives:
\begin{equation}
t(r) = r \theta \frac{h}{\sqrt{h^2 + (r \phi)^2}}
\end{equation}

The strut thicknesses at the inner and outer radii for each design are shown in Table~\ref{tab:parameters}. These geometric variations lead to different mechanical properties, which we characterize experimentally in Section~\ref{sec:mech_char}.

\subsection{Embedded Air Channel Design}

We implement a fluidic innervation strategy using air channels embedded within the helicoid struts. Details of the channel geometry and routing for the N6 design are shown in Fig.~\ref{fig:air_channels}. The channels have diameter 2 mm and are positioned 1.75 mm from the outer surface to maximize sensitivity to deformation while minimizing their impact on the structural integrity of the helicoid. 

Each segment is bounded by rigid end plates at the top and bottom (Fig.~\ref{fig:air_channels}, left). These end plates have three hook attachments for cable actuation, provide mounting surfaces for the custom PCBs, and enable robust mechanical connection between adjacent segments. Furthermore, since the end plates are rigid, their 6-DOF pose can be precisely defined and tracked, which facilitates kinematic modeling of the robot.

\subsection{Helicoid Fabrication}

The helicoid segments are fabricated using vision-controlled jetting~\cite{buchner2023vision}, a multi-material 3D printing technique that simultaneously deposits both soft and rigid materials in a single build. The compliant helicoid structure is printed from a thiol-ene polyurethane-based elastomer (TEPU 50A, Inkbit Corp.), while the segment end plates are printed from a tough epoxy (Titan Tough Epoxy 85, Inkbit Corp.). This method prints the structural helicoid geometry, rigid end plates, and embedded air channels in one fabrication step, which eliminates the need for complex assembly.
The printing process uses a dissolvable wax as support material. After printing, segments undergo post-processing to dissolve the wax support. This leaves a fully functional segment ready for integration into the robot assembly.

\subsection{PCB and Readout Electronics}

\begin{figure}
  \centering
  \includegraphics[width = 1.\columnwidth]{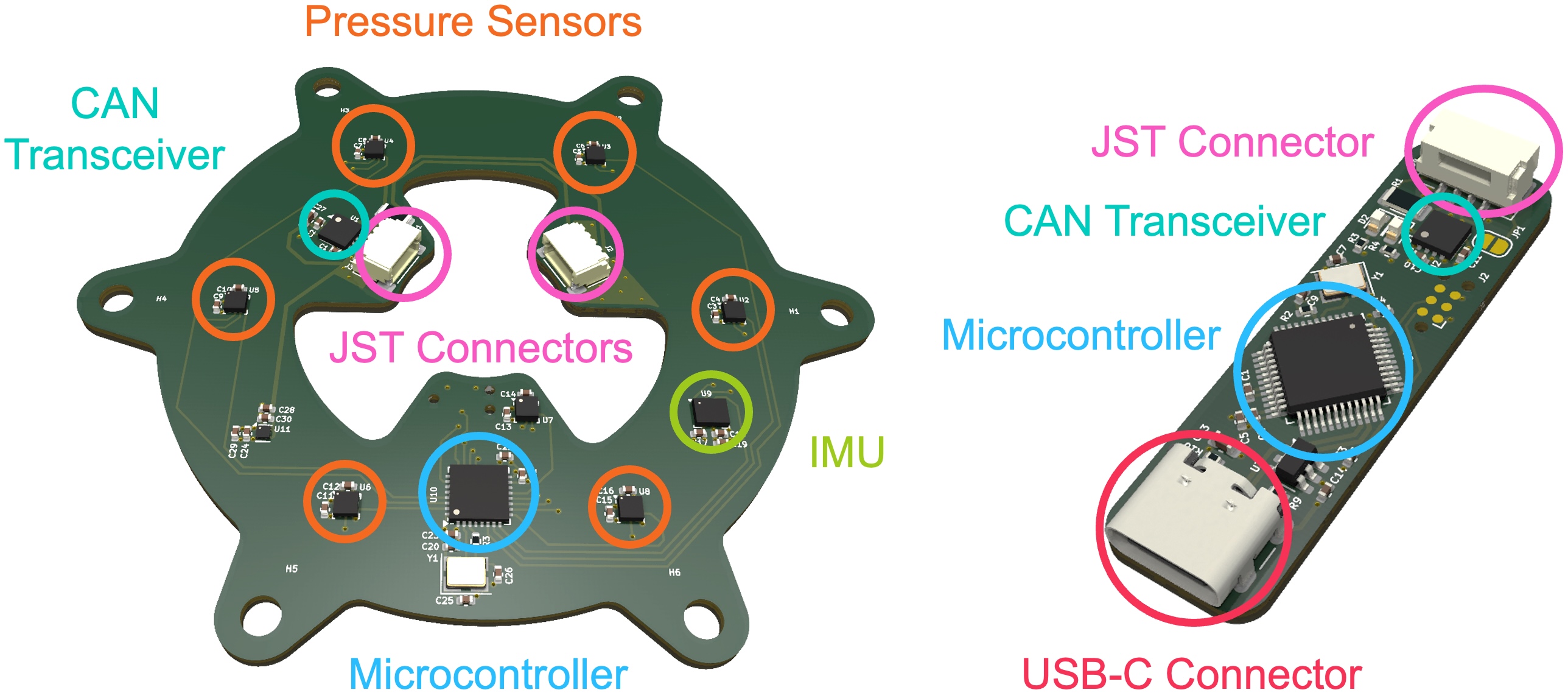}
  \caption{Left: Custom PCB with integrated pressure sensors and IMU. Right: CAN-USB adapter for interfacing with a host computer.}
  \label{fig:pcb}
  \vspace{-4mm}
\end{figure}

\begin{figure*}[t!]
  \centering
  \includegraphics[width = 1.\textwidth]{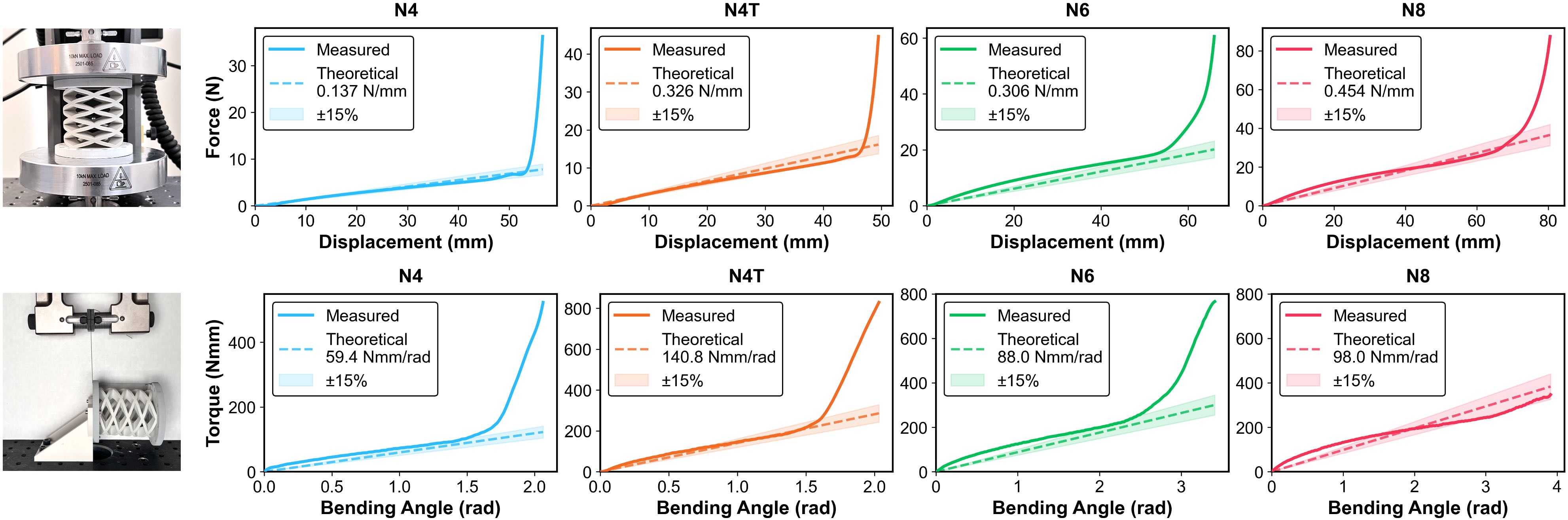}
  \caption{Mechanical characterization results. Top: axial stiffness and Bottom: bending stiffness for all four helicoid designs (N4, N4T, N6, N8) compared to theoretical predictions.}
  \label{fig:mechanical_char}
  \vspace{-4mm}
\end{figure*}

Each segment is equipped with a custom PCB (Fig.~\ref{fig:pcb}, left) containing six MEMS pressure sensors (LPS22HH, STMicroelectronics NV), a six-axis IMU (ICM-42688-P, InvenSense Inc.), and an onboard microcontroller. The pressure sensors provide absolute pressure measurements in the range 26--126 kPa with 0.02 Pa resolution at a maximum sampling rate of 200 Hz. The IMU provides 3-axis accelerometer and gyroscope data at up to 32 kHz. In practice, we sample all sensors at 200 Hz to balance temporal resolution with communication bandwidth constraints. The microcontroller reads sensor data and transmits it over a CAN bus interface. The PCBs can be daisy-chained, which eliminates the need for individual connections from each segment to the host computer and significantly reduces cable clutter inside the segments for multi-segment assemblies.
A custom USB-CAN adapter board (Fig.~\ref{fig:pcb}, right) receives CAN bus signals and converts them to USB for transmission to the host PC.

The air channel outlets at each segment end interface directly with the pressure sensors on the PCB. The soft elastomer material surrounding the channel outlets is shaped to create an airtight seal when compressed against the PCB-mounted sensors. The PCBs are bonded to the segments using epoxy adhesive.

\section{Experimental Evaluation}

\subsection{Mechanical Characterization}
\label{sec:mech_char}

Axial and bending stiffness are key performance metrics for trimmed helicoid structures, as they determine load-bearing capacity and compliance under actuation. We characterized both properties using an Instron universal testing machine under quasi-static loading. For axial stiffness, we applied compressive loads along the segment's central axis. For bending stiffness, we applied a moment at the segment end using a cable pulled by the Instron. The experimental setups are shown in the first column of Fig.~\ref{fig:mechanical_char}. We fabricated and tested all four helicoid designs (N4, N4T, N6, N8) from Table~\ref{tab:parameters} and compared measured stiffness values to theoretical predictions from beam theory~\cite{patterson2025design}.

To predict the mechanical behavior of our helicoid designs, we use the analytical model from~\cite{patterson2025design}. Each helical strut is modeled as a beam with a simplified rectangular cross-section of width $w$ and thickness $t_m=t(r_m)$ evaluated at the mid-radius $r_m = (r_i + r_o)/2$. The strut length is $L_m = L(r_m) = \sqrt{h^2 + (r_m \phi)^2}$.

The axial stiffness is derived from the textbook solution for transverse deflection of an Euler-Bernoulli beam:
\begin{equation}
k_{\text{ax}} = \frac{12 E I}{L_m^3 \cos^2(\alpha_m)}
\end{equation}
where $E$ is the Young's modulus (datasheet value of 2.48 MPa for the soft TEPU 50A material), $I$ is the second moment of area of the strut cross-section, and $\alpha_m = \alpha(r_m)$ is the helix angle at the mid-radius.
For a rectangular cross-section with width $w$ and thickness $t_m$, the relevant second moment of area is $I = \frac{w t_m^3}{12}$.

For bending stiffness, we model the helicoid as an equivalent hollow cylinder with the same bending behavior. The uncorrected bending stiffness is related to the axial stiffness by $k_{\text{bend,uncorr}} = k_{\text{ax}} \frac{I_c}{A}$, 
where $I_c = \pi r_m^4 / 4$ is the second moment of area of a circular cross-section at the mid-radius, and $A = \pi r_m^2$ is the cross-sectional area. 
However, this simple model underestimates the bending stiffness. Following~\cite{patterson2025design}, we apply an empirical correction factor of $\frac{9 r_m}{2 N h}$ derived from the Smalley equation for wave springs.
The final corrected bending stiffness is:
\begin{equation}
k_{\text{bend}} = k_{\text{ax}} \frac{I_c}{A} \frac{9 r_m}{2 N h}
\end{equation}

These theoretical predictions are plotted against the measured values in Fig.~\ref{fig:mechanical_char}. The force-displacement curves initially follow the predicted stiffness in a near-linear fashion, but then tick up sharply as the struts start to be in contact with each other. We find the predicted stiffness values to be in good agreement with the measured values, within $\pm$15\% for both axial and bending stiffness. Later when the struts undergo large deformations and start to be in contact with each other, the measured stiffness values deviate significantly from the predicted values.
As expected, both axial and bending stiffness increase with the number of helicoids $N$, due to the increased density of struts. Also, as expected, the N4T design has higher stiffness than the N4 design, due to the thicker struts.
Overall, this validates the analytical model and the proposed correction factor.

\subsection{Sensor Characterization}
\label{sec:sensor_char}

\begin{figure}
  \centering
  \includegraphics[width = 1.\columnwidth]{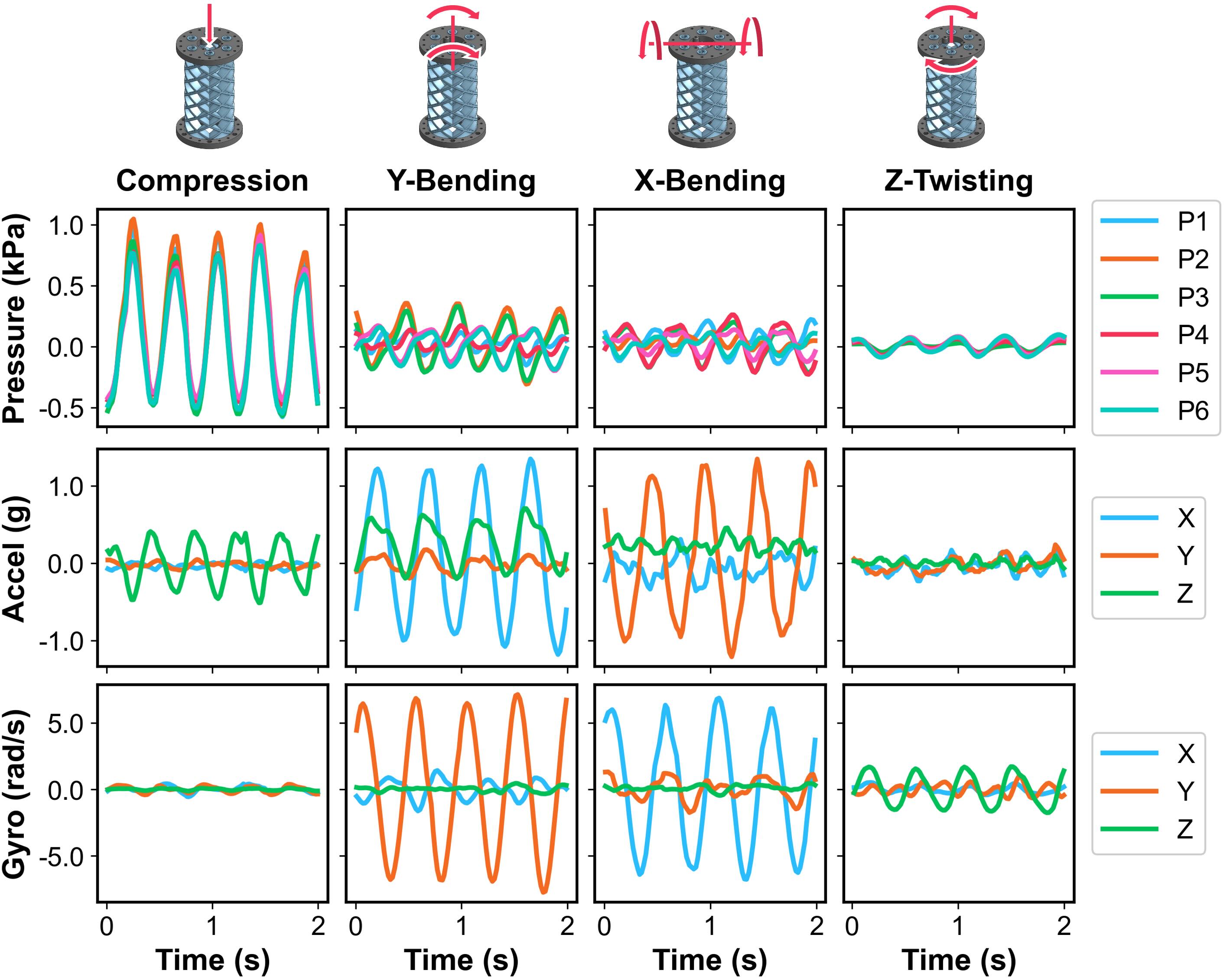}
  \caption{Sensor response characterization: pressure and IMU readings from the embedded air channels as the segment is subjected to different loading conditions.}
  \label{fig:sensor_char}
  \vspace{-4mm}
\end{figure}

We characterized the sensor response of the N6 design by subjecting a single segment to four distinct loading conditions: axial compression, bending in the 
east-west direction, bending in the north-south direction, and torsion about 
the central axis. The sensor data is low-pass filtered and then zeroed to measure changes relative to the undeformed state of the segment. Figure~\ref{fig:sensor_char} shows representative time series data from the six pressure sensors and the IMU during these deformations. Each loading mode produces a distinct and repeatable signature across the sensor array, confirming that the sensors provide sufficient information to distinguish between fundamental deformation modes. 

The sensing principle follows established fluidic innervation approaches~\cite{truby2022fluidic, zhang2024embedded, zhang2025fluidically}: deformations that compress the air channels increase pressure, while those that expand the channels decrease pressure relative to ambient. During bending, channels on the compressed side show increased pressure while those on the tensioned side show decreased pressure. Tactile interactions and contact forces similarly produce localized pressure changes in the corresponding channels. Detailed characterizations of fluidic innervation sensing, including cyclic repeatability and material robustness, are reported in~\cite{truby2022fluidic}.

\section{14-DoF Robot Arm}

Based on the mechanical characterization results, we selected the N6 design for the complete robot arm system. This design provides a favorable balance between compliance for actuation and structural rigidity for load bearing. N4T and N6 have similar axial stiffness, but N6 has lower bending stiffness, which makes them easier to actuate. Also N6 segments are 50\% longer than N4T segments, which allows for a larger workspace for the same number of segments, or requires fewer segments to achieve the same workspace.

\subsection{Gripper}

We designed a single-DoF cable-driven, tulip-inspired soft gripper that attaches to the distal end of the robot arm (Fig.~\ref{fig:robot_assembly}~Left and Bottom Right). The gripper features four embedded air channels for tactile feedback, with a hardware interface compatible with the same PCBs used for the helicoid segments. The deformable part is 3D printed using vision-controlled jetting from the soft TEPU 50A elastomer, while the mounting plate and cable attachment hook use the same rigid epoxy as the segment end plates.

\subsection{Motor Base and Assembly}

\begin{figure}
  \centering
  \includegraphics[width = .9\columnwidth]{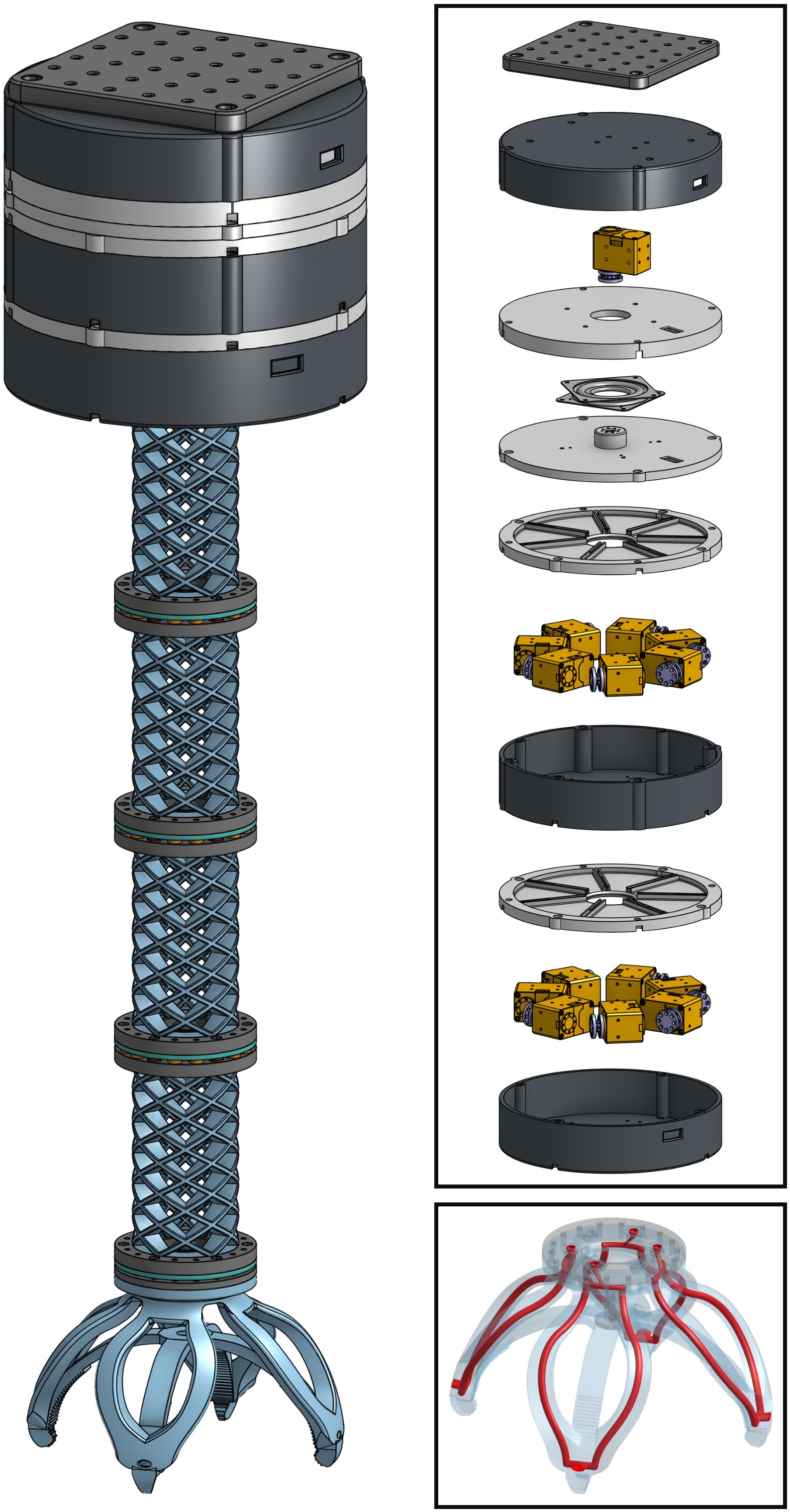}
  \caption{Left: Full arm assembly. Top Right: Exploded view of motor base assembly showing motor arrangement, cable guide plates, and mounting plates. Bottom Right: Transparent view of gripper revealing the embedded air channels.}
  \label{fig:robot_assembly}
  \vspace{-4mm}
\end{figure}

The complete robot arm consists of four N6 helicoid segments stacked to create a 12-DoF system, with each segment actuated by three cables driven by servo motors (Dynamixel XM430-W350-R, ROBOTIS) housed in a modular motor base. Fig.~\ref{fig:robot_assembly} shows the complete arm assembly and an exploded view of the motor base. 

The motor base features a modular design where each layer houses up to seven motors, with cables wound on 3D-printed spools. Cable guide plates are integrated at the top of each layer to route the cables to their respective segments. An additional motor enables rotation of the entire robot about its central axis, supported by an axial bearing. An optical mounting plate (Newport Corp.) at the top facilitates positioning and portability of the arm. All structural components (motor layers, cable guides, and mounting plates) are 3D-printed from PLA using a desktop FDM printer (P1S, Bambu Lab).

The complete system provides 14 DoF: one for base rotation, 12 for the four helicoid segments (three DoF each), and one for the gripper. The total length from motor base to gripper tip is approximately 72 cm. The robot is mounted within a workspace cage of 2'$\times$2'$\times$3' constructed from aluminum extrusion profiles, which delineates the workspace of the robot.

\subsection{Experiments}

We conducted a series of experiments to demonstrate the mechanical capabilities of the complete robot platform. The trajectory tracking and grasping experiments use open-loop control based on the kinematic model to validate the platform's mechanical performance. While the segment sensors were characterized at the individual segment level (Section~\ref{sec:sensor_char}), developing proprioceptive control algorithms that leverage these sensors for closed-loop control of the full arm is beyond the scope of this work. However, we demonstrate proof-of-concept tactile sensing using the gripper's embedded sensors for object stiffness detection.

\begin{figure}
  \centering
  \includegraphics[width = 1.\columnwidth]{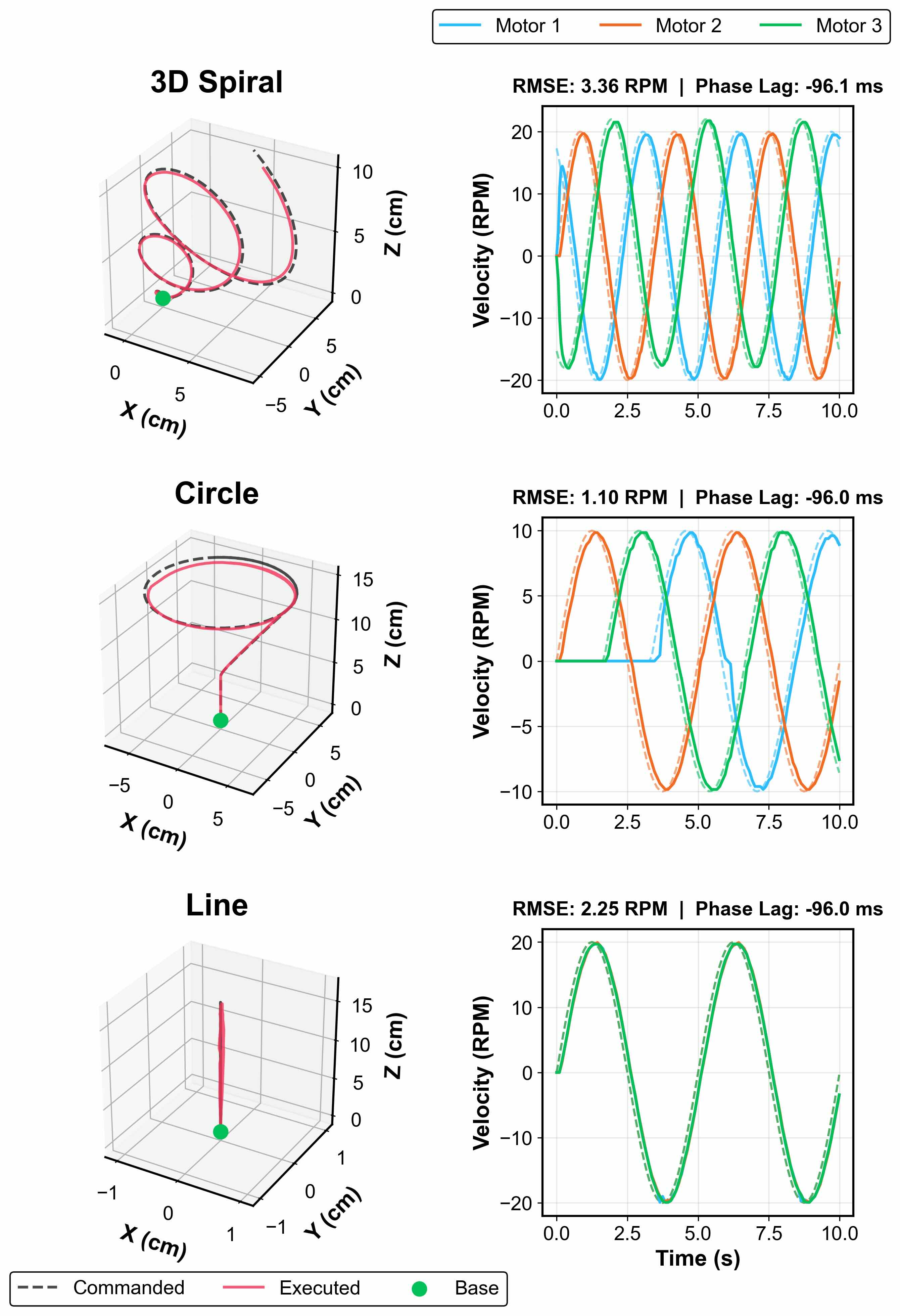}
  \caption{Trajectory tracking results for three different paths. Left column: 3D task space trajectories showing commanded (black dashed) vs. executed (red solid) paths reconstructed from motor encoders via forward kinematics. Right column: Velocity tracking for the three motors, with performance metrics (RMSE and phase lag) shown above each plot.}
  \label{fig:trajectory_tracking}
  \vspace{-4mm}
\end{figure} 

\subsubsection{Trajectory Tracking}

\begin{figure}
  \centering
  \includegraphics[width = 1.\columnwidth]{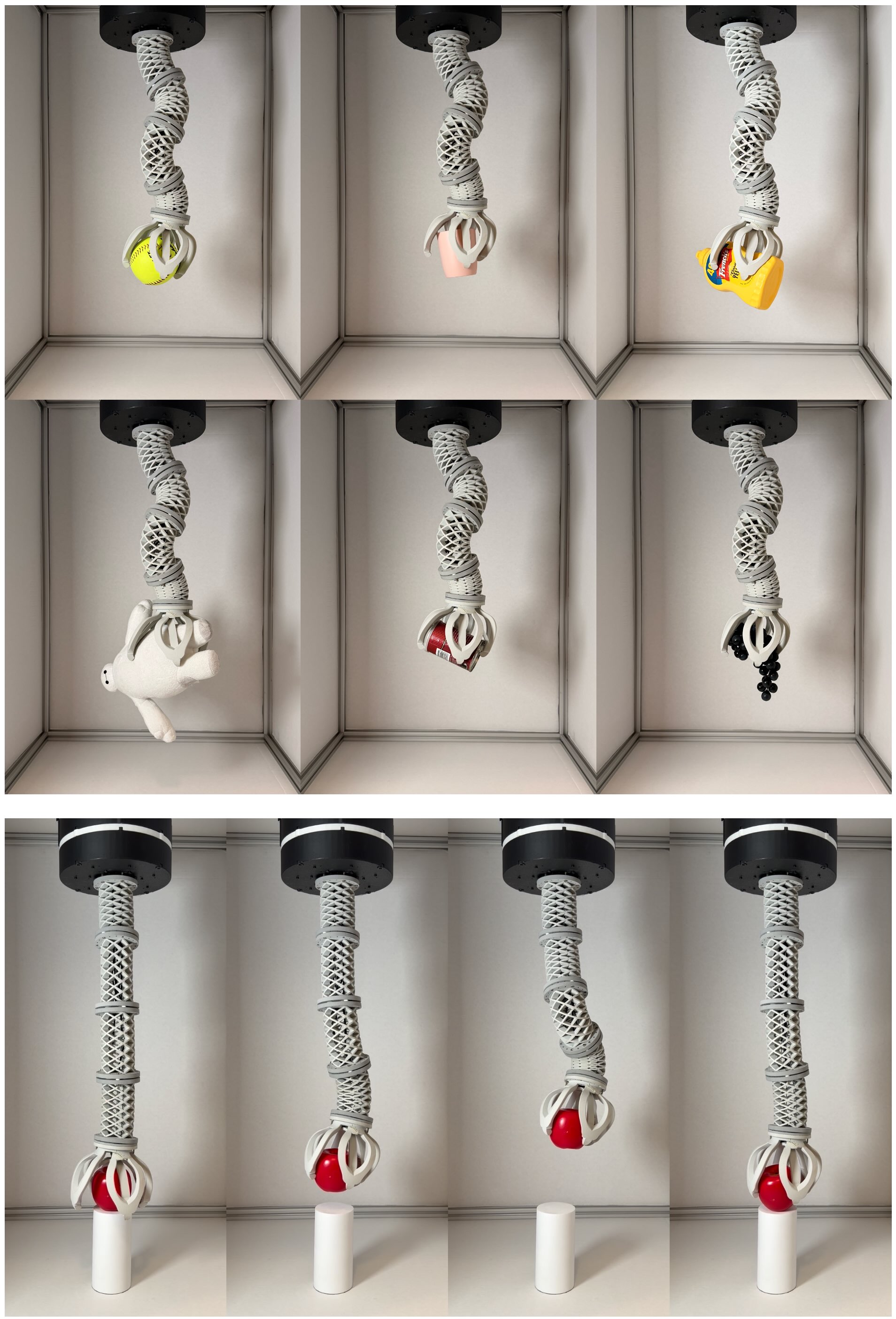}
  \caption{Object grasping. Top: The gripper successfully adapts to objects of various shapes, sizes, and stiffnesses. Bottom: Sequential images showing the robot picking up an apple from a pedestal and placing it back down.}
  \label{fig:grasping}
  \vspace{-4mm}
\end{figure}

We developed a piecewise constant curvature kinematic model that maps motor positions to the pose of the robot segments. This model assumes that each segment deforms with constant curvature, a common assumption for continuum robots~\cite{webster2010design}. The forward kinematics model takes cable lengths (determined by motor positions) as inputs and outputs the 3D pose of each segment.

To evaluate the robot's trajectory tracking capabilities, specifically when the motors are under load, we commanded three distinct Cartesian space trajectories in open-loop control: a vertical line, a horizontal circle, and a 3D spiral with varying elevation. For each trajectory, we computed the inverse kinematics to determine the required motor velocities, commanded these velocities to the motors, and recorded both the commanded and executed velocities.

Figure~\ref{fig:trajectory_tracking} shows the results for all three trajectories. The left column displays the 3D task space trajectories, comparing the commanded path (black dashed line) with the executed path reconstructed from motor encoder data via forward kinematics (red solid line). The right column shows the velocity tracking performance for the active motors, with commanded velocities shown as dashed lines and executed velocities as solid lines. The velocity tracking metrics (RMSE and phase lag) quantify the control accuracy at the motor level.

\subsubsection{Object Grasping}
We demonstrated the robot's versatility in manipulating diverse objects using the gripper. The system successfully grasped objects of varying shapes, sizes, and material properties, as shown in Fig.~\ref{fig:grasping} and supplementary video clip 1. The two top rows demonstrate the gripper's adaptability to objects of different geometries, including cylindrical, spherical, and irregularly shaped items. The bottom row shows a sequence from a video of a pick-and-place task where the robot grasps an apple from a pedestal, lifts it to some predetermined pose, and returns it to the original position (supplementary video clip 2). The compliant nature of the gripper enables secure grasping without requiring precise force control or complex grasp planning.

\begin{figure}
  \centering
  \includegraphics[width = 1.\columnwidth]{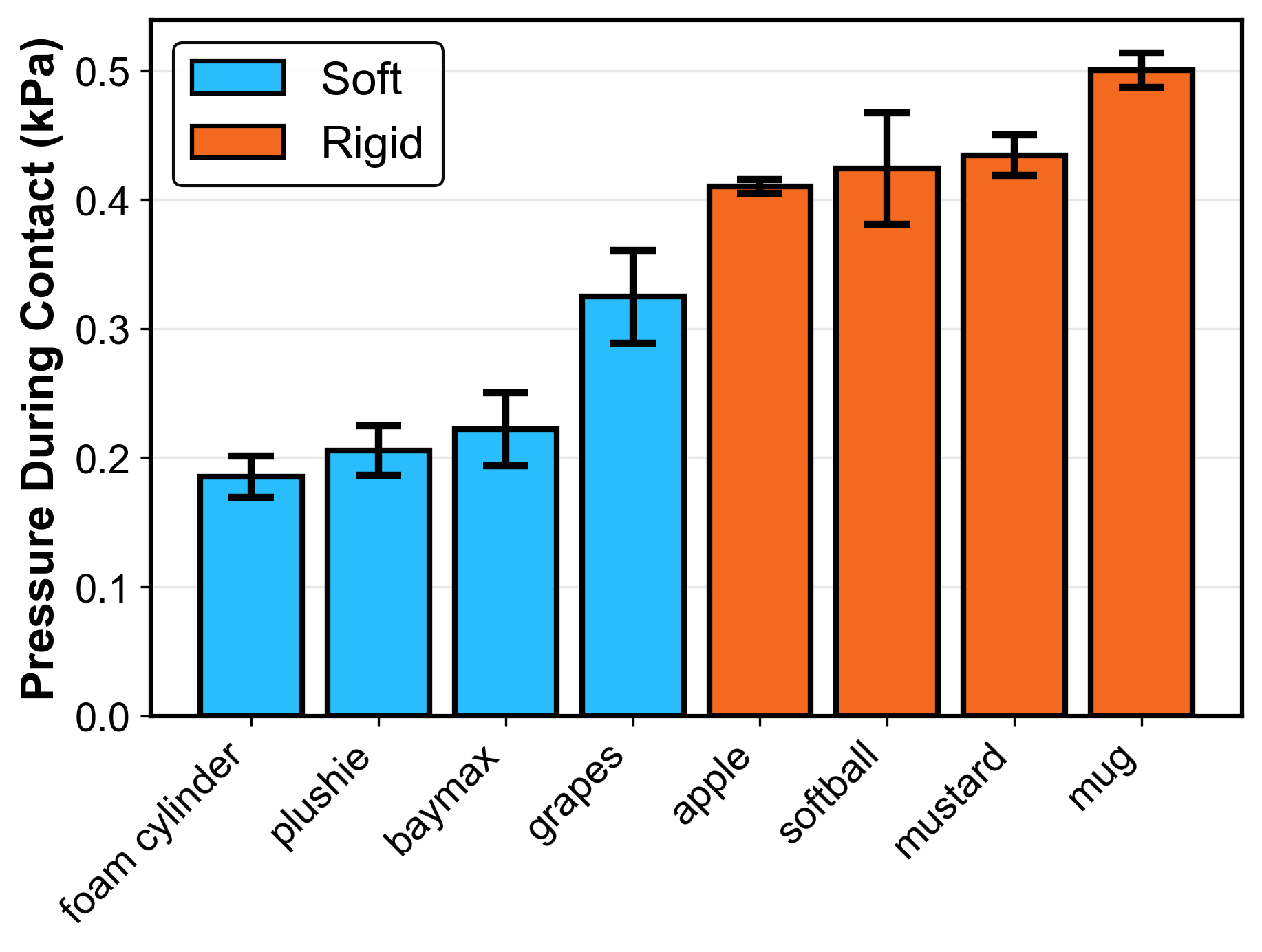}
  \caption{Object stiffness detection: Average pressure in the air channels as the robot is in contact with different objects. Error bars show standard deviation.}
  \label{fig:object_stiffness}
  \vspace{-4mm}
\end{figure}

\subsubsection{Object Stiffness Detection}
We explored the use of embedded pressure sensors for inferring object stiffness during grasping and manipulation. When the gripper contacts an object and applies a grasping force, the pressure readings in the air channels provide information about the mechanical interaction. Our hypothesis is that contact with compliant objects results in lower pressure readings in the gripper's air channels compared to rigid objects, as the soft material deforms to accommodate the gripper's geometry, which reduces the local contact pressures.

Figure~\ref{fig:object_stiffness} shows the average pressure measured across the gripper's air channels when grasping objects of varying stiffness with minimal grasping force required to hold the object. The results demonstrate a measurable and consistent difference between soft and rigid objects in our experiments. While this relationship is influenced by multiple factors, including object weight, surface friction, contact area, and grasp pose, the pressure magnitude nonetheless provides a useful empirical indicator of object compliance. This suggests that the embedded fluidic sensors can serve as a simple tactile feedback mechanism for distinguishing between broad categories of object stiffness.

\section{Conclusion}
This paper presents a fabrication method for sensorized soft continuum robot segments combining trimmed helicoid structures with fluidic innervation. Using vision-controlled jetting, we fabricated multi-material segments with integrated air channels interfacing with custom PCBs housing miniature pressure sensors and IMUs. Mechanical characterization of four helicoid designs identified design tradeoffs, and sensor characterization validated the response to fundamental deformation modes. A meter-scale, 14-DoF cable-driven robot arm demonstrated the platform's scalability and mechanical capabilities, including open-loop trajectory tracking and object grasping. Proof-of-concept tactile sensing using the gripper's embedded sensors successfully demonstrated object stiffness detection. These results validate the feasibility of this fabrication approach for creating sensorized, architected soft structures at scale.

Current limitations include oscillations during dynamic operation and sensor drift and hysteresis, which are challenges inherent to the soft material rather than the sensing approach itself. Future work will address these limitations through stiffer material formulations or compensation algorithms. Inspired by recent advances in interpreting fluidic sensor data with machine learning~\cite{zhang2023machine,truby2022fluidic}, we aim to develop a model to map the pressure readings to the segment's configuration parameters, which would enable closed-loop trajectory tracking. Further directions include contact detection for safe interaction with unstructured environments, alternative channel routing and sensor multiplexing to improve information density, and multi-modal sensor fusion (pressure and IMU) for behaviors such as obstacle avoidance and tactile exploration.






\section*{Acknowledgments}

The authors thank Kiwan Wong for help with experiments and Liong Ma for help with PCB firmware. This work was supported by the Singapore-MIT Alliance for Research and Technology (SMART) Mens, Manus, and Machina program and the Gwangju Institute of Science and Technology.


\bibliographystyle{IEEEtran}
\bibliography{library}

\end{document}